\documentclass[letterpaper]{article} % DO NOT CHANGE THIS
\usepackage[preprint]{aaai2027}
\usepackage[hyphens]{url}  % DO NOT CHANGE THIS
\usepackage{graphicx} % DO NOT CHANGE THIS
\usepackage{natbib}  % DO NOT CHANGE THIS AND DO NOT ADD ANY OPTIONS TO IT
\usepackage{caption} % DO NOT CHANGE THIS AND DO NOT ADD ANY OPTIONS TO IT
\usepackage{algorithm}
\usepackage{algorithmic}
\usepackage{multirow}
\usepackage{array}
\usepackage{amsfonts}
\usepackage{comment}
\usepackage{amsmath}
\usepackage{array}
\newcolumntype{L}{>{\raggedright\arraybackslash}p{2cm}}
\usepackage{newfloat}
\usepackage{listings}
\DeclareCaptionStyle{ruled}{labelfont=normalfont,labelsep=colon,strut=off} % DO NOT CHANGE THIS
\floatstyle{ruled}
\newfloat{listing}{tb}{lst}{}
\floatname{listing}{Listing}

\usepackage{booktabs}

\title{Rethinking Modality Reliability in Multimodal Sentiment Analysis with Incomplete Observations}
\author{
    Chunlei Meng\textsuperscript{\rm 1}\thanks{clmeng23@m.fudan.edu.cn}\\
    Jacqueline J. Pang\textsuperscript{\rm 2}, Pengbin Feng\textsuperscript{\rm 3}, Zhenyu Yu\textsuperscript{\rm 4} \\
    Chun Ouyang\textsuperscript{\rm 1}\corresponding, Zhongxue Gan\textsuperscript{\rm 1}   }
\affiliations{
    \textsuperscript{\rm 1}The College of Intelligent Robotics and Advanced Manufacturing (CIRAM), Fudan University\\

    \textsuperscript{\rm 2}Cornell University, 
    \textsuperscript{\rm 3}University of Southern California (USC)\\
    \textsuperscript{\rm 4}The College of Computer Science and Artificial Intelligence, Fudan University\\
    
}

\begin{document}

\maketitle

\begin{abstract}
Multimodal Sentiment Analysis (MSA) integrates text, audio, and vision to infer human affect, yet real-world multimodal observations are often incomplete. Existing methods for incomplete-observation MSA mainly follow two paradigms. Reconstruction-based methods recover missing information from observed modalities, while joint-representation methods learn directly from incomplete inputs. Although effective, these methods usually treat modality reliability only implicitly within representation learning or fusion design rather than modeling it explicitly. We argue that modality reliability is a central variable in incomplete-observation settings. Failure to model it explicitly gives rise to two related issues. The first is reliability mismatch, in which the affective evidence retained by each modality varies across samples and missing rates. The second is reliability propagation bias, in which messages from degraded modalities may adversely affect cross-modal interaction and predictive performance. To address these issues, we propose MRCF, a Modality Reliability-Calibrated Framework for MSA with incomplete observations. MRCF contains a Reliability-Aware Branch that estimates sample-specific modality reliability from intramodal quality cues and cross-modal semantic consistency, a Reliability-Guided Interaction Branch that uses the estimated scores to modulate cross-modal information flow, and a Reliability-Calibrated Fusion Module that integrates reliability and semantic cues for final prediction. Experiments on CMU-MOSI, CMU-MOSEI, and CH-SIMS show that MRCF achieves strong performance under standard incomplete-observation protocols. Further analyses provide evidence that explicit reliability modeling helps mitigate reliability mismatch and reliability propagation bias during interaction and fusion.
\end{abstract}

\begin{comment}
\begin{figure}[!t]
    \centering
    \includegraphics[width=1\linewidth]{figures/motivation-v1.png}
    \caption{Preliminary diagnosis of modality reliability under incomplete observations and its effect on prediction. (a) As the missing rate increases, modality reliability decreases at different rates; audio and vision alternate as the least reliable modality across missing stages. (b) The annotations report $\Delta\mathrm{F1}=\mathrm{F1}_{\rm drop}-\mathrm{F1}_{\rm all}$. Negative values indicate that dropping the selected modality hurts performance, whereas positive values indicate that removing it improves performance and therefore suggest negative transfer from its inclusion.}
    \label{fig:motivation}
\end{figure}    
\end{comment}

\begin{figure}[!t]
    \centering
    \includegraphics[width=1\linewidth]{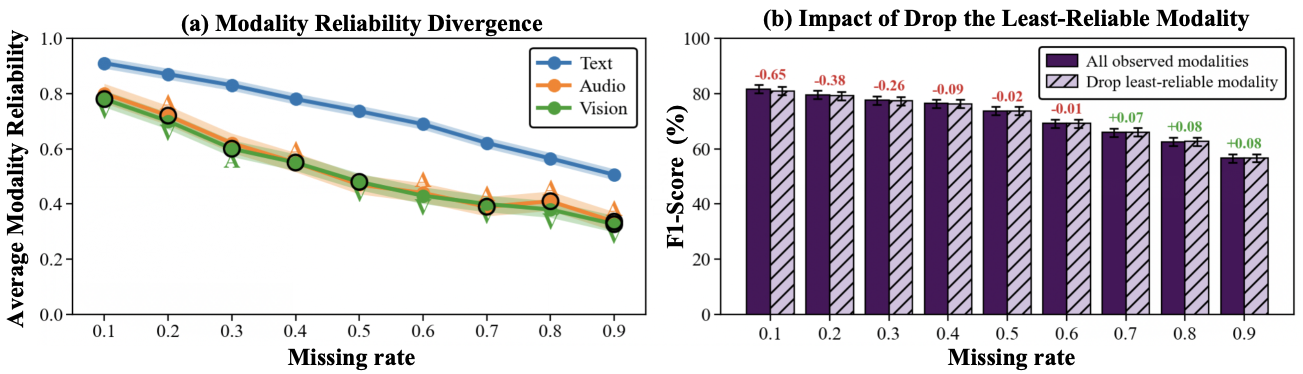}
    \caption{Diagnosis of modality reliability under incomplete observations.
(a) Reliability declines differently across modalities as the missing rate increases, so the least-reliable modality changes across stages.
(b) Dropping the least-reliable modality yields
$\Delta\mathrm{F1}=\mathrm{F1}_{\mathrm{drop}}-\mathrm{F1}_{\mathrm{all}}$;
negative values indicate performance degradation, whereas positive values indicate improvement and suggest negative transfer.}
    \label{fig:motivation}
\end{figure}

\begin{figure*}[!t]
    \centering
    \includegraphics[width=0.9\linewidth]{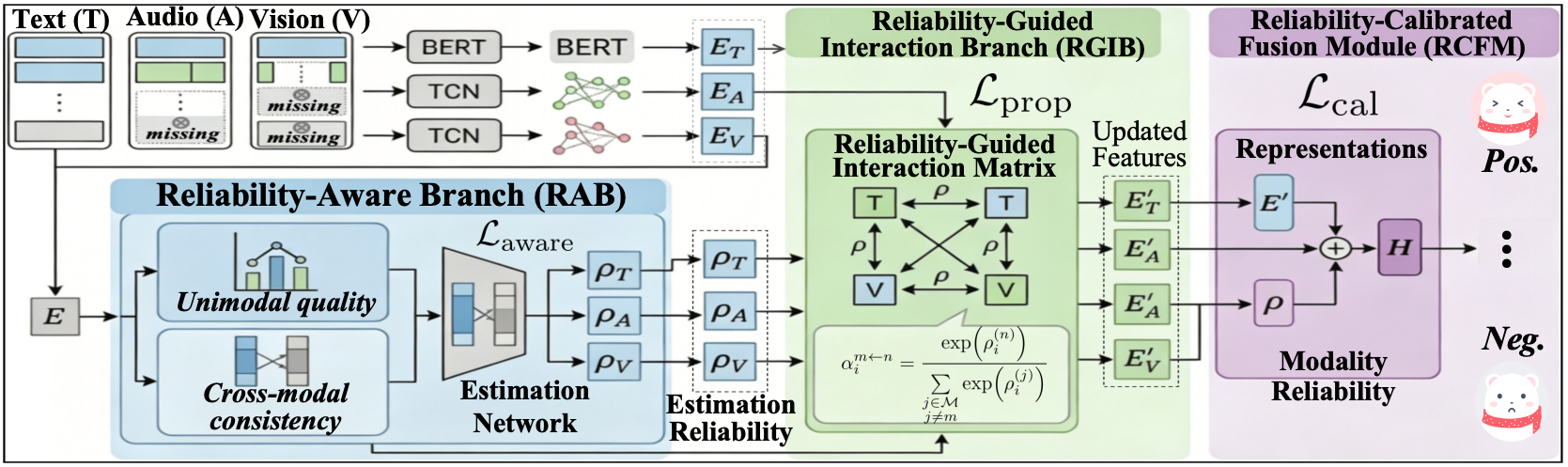}
\caption{Overview of MRCF. Given incomplete inputs, modality-specific encoders first extract unimodal features. The Reliability-Aware Branch estimates modality reliability using intramodal quality and cross-modal consistency cues. Guided by these scores, the Reliability-Guided Interaction Branch modulates cross-modal message passing according to estimated reliability. Finally, the Reliability-Calibrated Fusion Module integrates semantic and reliability cues for sentiment prediction. Paired complete observations are used only during training for proxy supervision and fusion calibration}
\label{fig:MRCF}
\end{figure*}

\section{Introduction}

Multimodal Sentiment Analysis (MSA) aims to infer human sentiment by integrating complementary cues from text, audio, and vision~\cite{CLCR,TSDA,GCL,MGJR}. Its effectiveness relies on the fact that different modalities provide heterogeneous yet mutually informative evidence for affect understanding~\cite{misa,DBR,TSD}. In real-world applications, multimodal observations are often incomplete due to sensor failures, occlusion, noise, privacy constraints, and transcription errors~\cite{LNLN,PRMF,AHRA}. Under such conditions, models must infer sentiment reliably from partial evidence~\cite{ROSA}, making incomplete-observation MSA increasingly important in practice.

Existing studies mainly address this problem through two paradigms: reconstruction-based learning and joint-representation learning. These paradigms differ primarily in whether missing information is explicitly recovered before downstream prediction. Reconstruction-based methods attempt to recover missing information from observed modalities before performing downstream prediction~\cite{NIAT}. Representative methods include TFR-Net~\cite{TFR-Net}, which employs attention-based extractors and a reconstruction module to recover missing semantics in unaligned multimodal sequences, and DAR~\cite{DRA}, which decouples modality-common and modality-specific semantics and reconstructs them before fusion. Joint-representation methods learn robust multimodal representations directly from incomplete inputs via distillation, dominant-modality preservation, proxy modeling, or fusion redesign~\cite{HKD-MER}. CoKD~\cite{CorrKD} uses correlation-decoupled distillation for uncertain missing patterns, LNLN~\cite{LNLN} preserves the dominant language modality, and P-RMF~\cite{PRMF} models proxy modalities and uncertainty in a Gaussian latent space. Although designed for complete observations, ALMT~\cite{ALMT} offers related insight through language-guided suppression of disruptive auxiliary cues. See Appendix~A for details.

Despite these advances, existing methods do not explicitly and systematically model modality reliability under incomplete observations. Instead, their focus on missing-information recovery or representation fusion leaves reliability implicit rather than treating it as a sample-specific variable. Fig.~\ref{fig:motivation} provides two diagnostic analyses of this issue. In Fig.~\ref{fig:motivation}(a), we use the agreement between a modality-specific prediction and the corresponding full-observation prediction at each missing rate as a proxy for modality reliability. The resulting curves show that this reliability proxy is neither uniform across modalities nor static across missing rates. As the missing rate increases, the proxy scores of different modalities decrease at different rates. In particular, while text remains the highest-scoring modality overall, audio and vision alternate as the lowest-scoring modality at different missing stages, revealing a clear reliability mismatch across modalities. This indicates that the affective evidence retained under incomplete observations varies dynamically across modalities and corruption levels, whereas current methods do not generally expose this variation as an explicit sample-specific quantity. Figure~\ref{fig:motivation}(b) further examines the predictive effect of removing the lowest-scoring modality at each missing rate. The comparison shows a systematic transition as missingness increases. At low missing rates, dropping the lowest-scoring modality leads to a noticeable performance decrease, suggesting that even a relatively low-scoring modality may retain weak but useful complementary cues. As the missing rate increases further, the performance gap gradually shrinks and eventually becomes slightly positive, indicating that evidence from the lowest-scoring modality becomes increasingly likely to disrupt interaction and fusion rather than provide useful support. This pattern is consistent with reliability propagation bias: errors introduced by degraded evidence may spread through cross-modal interaction and adversely affect the joint representation. This interpretation is also consistent with several recent findings. LNLN~\cite{LNLN} identifies temporal information corruption and modality laziness under random missingness, P-RMF~\cite{PRMF} highlights unequal modality contributions, cross-modal discrepancies, and uncertainty, and CMAD~\cite{CMAD} reveals that uncontrolled interaction causes representation inconsistency and optimization instability. These findings motivate explicitly modeling sample-specific modality reliability under incomplete observations.

Motivated by these, we propose the Modality Reliability-Calibrated Framework (MRCF) for MSA with incomplete observations. MRCF contains two complementary branches and a reliability-calibrated fusion module. First, a Reliability-Aware Branch explicitly estimates sample-specific modality reliability by jointly modeling intramodal quality cues and cross-modal semantic consistency and is designed to address reliability mismatch. Second, a Reliability-Guided Interaction Branch uses the estimated reliability to regulate cross-modal information flow, assigning relatively greater influence to messages from higher-scoring modalities and reducing the influence of lower-scoring ones, thereby targeting reliability propagation bias during interaction. Finally, a Reliability-Calibrated Fusion Module integrates semantic and reliability cues for final prediction and uses complete observations as training-time anchors for proxy reliability supervision and fusion calibration. In contrast to methods that focus mainly on processing incomplete inputs, our framework emphasizes estimating modality reliability, using it during interaction, and calibrating the final fusion decision. 

The contributions are: \textbf{1.} We characterize dynamic modality reliability in incomplete-observation MSA and propose the Modality Reliability-Calibrated Framework (MRCF) to address reliability mismatch and propagation bias. \textbf{2.} We design a Reliability-Aware Branch (RAB) to estimate sample-specific modality reliability from intra- and cross-modal cues, and a Reliability-Guided Interaction Branch (RGIB) to suppress unreliable cross-modal messages. \textbf{3.} We introduce a Reliability-Calibrated Fusion Module (RCFM) that jointly models semantic and reliability information and uses a complete-observation semantic anchor to calibrate fusion, yielding stable representations and improved prediction.

\section{Methodology}

\subsection{Overview}

We study MSA with incomplete observations. Given an input sample, one or more modalities may be partially missing, making the remaining affective evidence both incomplete and unevenly reliable across modalities. To reduce the influence of degraded evidence on cross-modal interaction, we propose the Modality Reliability-Calibrated Framework (MRCF). As shown in Fig.~\ref{fig:MRCF}, MRCF contains three tightly coupled components.
First, a Reliability-Aware Branch estimates sample-specific modality reliability from intramodal quality cues and cross-modal semantic consistency.
Second, a Reliability-Guided Interaction Branch uses the estimated reliability to modulate cross-modal information flow, assigning relatively greater influence to messages from higher-scoring modalities.
Third, a Reliability-Calibrated Fusion Module integrates semantic and reliability cues for final prediction and aligns incomplete-observation fusion with a complete-observation semantic anchor during training. A key property of MRCF is that modality reliability is treated as an explicitly estimated quantity rather than an implicit by-product of feature learning.
Accordingly, the framework is optimized for two coupled goals: estimating modality reliability under incomplete observations and using it to reduce reliability propagation bias during interaction and fusion.

\subsection{Feature Extraction}

Let $\mathcal{M}=\{t,a,v\}$ denote the modality set, where $t$, $a$, and $v$ represent text, audio, and vision, respectively. Given a training set $\mathcal{D}=\{(\mathbf{X}_i,y_i)\}_{i=1}^{N}$, the complete multimodal input of the $i$-th sample is written as $\mathbf{X}_i=\{X_i^{(m)}\}_{m\in\mathcal{M}}$, where $X_i^{(m)}\in\mathbb{R}^{L_i^{(m)}\times d_m}$ is the modality-specific sequence, $L_i^{(m)}$ is the sequence length, and $d_m$ is the feature dimension. The sentiment label is $y_i\in\mathbb{R}$. Following the standard incomplete-observation protocol, we construct an incomplete counterpart $\widetilde{\mathbf{X}}_i=\{\widetilde{X}_i^{(m)}\}_{m\in\mathcal{M}}$ by applying $\widetilde{X}_i^{(m)}=\mathcal{C}\!\left(X_i^{(m)},B_i^{(m)}\right)$, where $B_i^{(m)}\in\{0,1\}^{L_i^{(m)}}$ is a binary observation mask, $B_{ij}^{(m)}=1$ denotes an observed temporal position, and $\mathcal{C}(\cdot)$ is the corruption operator. We define the observed ratio of modality $m$ as
\begin{comment}
\begin{equation}
\gamma_i^{(m)}=\frac{1}{L_i^{(m)}}\sum_{j=1}^{L_i^{(m)}} B_{ij}^{(m)}.
\end{equation}    
\end{comment}
$\gamma_i^{(m)}=\frac{1}{L_i^{(m)}}\sum_{j=1}^{L_i^{(m)}} B_{ij}^{(m)}.$
This input construction is standard and not our primary contribution. Unlike existing incomplete-observation MSA methods that directly reconstruct, distill, or fuse masked inputs, we explicitly model modality reliability as a sample-specific quantity. Since the affective evidence retained by each modality varies across samples and missing rates, the model should first assess what information remains and how reliable it is before cross-modal interaction.

For each modality $m\in\mathcal{M}$, a modality-specific encoder $E_m$ maps the incomplete input to a hidden sequence $\mathbf{H}_i^{(m)} = E_m\!\left(\widetilde{X}_i^{(m)}\right)$, with $\mathbf{H}_i^{(m)}\in\mathbb{R}^{L_i^{(m)}\times d}$, where $d$ is the shared hidden dimension. A pooling layer then produces a compact modality representation $\mathbf{h}_i^{(m)} = \mathrm{Pool}\!\left(\mathbf{H}_i^{(m)}\right)\in\mathbb{R}^{d}$. During training, we additionally encode the complete observations using the same encoders: $\mathbf{H}_{i,c}^{(m)} = E_m\!\left(X_i^{(m)}\right)$ and $\mathbf{h}_{i,c}^{(m)} = \mathrm{Pool}\!\left(\mathbf{H}_{i,c}^{(m)}\right)$. These complete-observation features are used only as training-time anchors for reliability supervision and fusion calibration, and are not required during inference.

\subsection{Reliability-Aware Branch}

The first component of MRCF estimates how trustworthy each modality is for the current sample under incomplete observations. This branch is motivated by the fact that modality reliability should not be treated as a static prior, because it depends jointly on what remains observable within the modality and whether the retained semantics agree with those of other modalities. To this end, we use two complementary cues.

\textbf{Intramodal quality cue.} We first characterize the internal quality of the observed modality itself. The pooled modality representation is combined with its observed ratio to obtain an intramodal quality feature $\mathbf{q}_i^{(m)} = \phi_q^{(m)}\!\left([\mathbf{h}_i^{(m)};\gamma_i^{(m)}]\right)\in\mathbb{R}^{d_q}$, where $\phi_q^{(m)}(\cdot)$ is a modality-specific projection network and $[\cdot;\cdot]$ denotes concatenation.

\textbf{Cross-modal semantic consistency cue.} We then evaluate whether modality $m$ remains semantically consistent with the other modalities. If it still preserves useful affective evidence, its representation should be more compatible with the representations derived from the others. We therefore define the semantic consistency score of modality $m$ as
\begin{equation}
s_i^{(m)}=
\frac{1}{|\mathcal{M}|-1}
\sum_{\substack{n\in\mathcal{M}\\ n\neq m}}
\cos\!\left(
W_s^{(m)}\mathbf{h}_i^{(m)},
W_s^{(n)}\mathbf{h}_i^{(n)}
\right),
\end{equation}
where $W_s^{(m)}\in\mathbb{R}^{d\times d}$ is a learnable projection matrix and $\cos(\cdot,\cdot)$ is cosine similarity.

\textbf{Reliability estimation.} Modality reliability is estimated by combining the two cues through a nonlinear mapping:
\begin{equation}
    \rho_i^{(m)}=
\sigma\!\left(
\mathbf{w}_r^\top
\tanh\!\left(
W_r[\mathbf{q}_i^{(m)};s_i^{(m)}]+\mathbf{b}_r
\right)
\right).
\end{equation}
Here $W_r\in\mathbb{R}^{d_r\times(d_q+1)}$ and $\mathbf{w}_r,\mathbf{b}_r\in\mathbb{R}^{d_r}$ are learnable parameters, and the sigmoid gives $\rho_i^{(m)}\in(0,1)$. A larger $\rho_i^{(m)}$ means that modality $m$ is estimated to provide more trustworthy affective evidence for sample $i$.

\textbf{Proxy reliability supervision.} Since complete observations are available during training, we construct a training-time proxy target from the agreement between the incomplete unimodal prediction and the complete-observation prediction. For modality $m$, we first obtain a unimodal prediction $\hat{y}_i^{(m)} = g_m\!\left(\mathbf{h}_i^{(m)}\right)$. The complete-observation prediction $\hat{y}_i^{c}$ is obtained from the complete-observation fusion anchor defined in Sec.~\ref{sec:rcfm}. We then define $\bar{\rho}_i^{(m)}=
\exp\!\left(
-\left|
\hat{y}_i^{(m)}-\operatorname{sg}\!\left(\hat{y}_i^{c}\right)
\right|
\right)$, where $\operatorname{sg}(\cdot)$ stops gradients through the complete-observation prediction. This target is a model-derived training proxy, not a ground-truth reliability label; it reflects how closely an incomplete unimodal prediction agrees with the complete-observation prediction.

The reliability regression loss is:
\begin{equation}
    \mathcal{L}_{\mathrm{rel}}=
\frac{1}{N|\mathcal{M}|}
\sum_{i=1}^{N}\sum_{m\in\mathcal{M}}
\mathrm{SL1}\!\left(\rho_i^{(m)},\bar{\rho}_i^{(m)}\right),
\end{equation}
where $\mathrm{SL1}(\cdot,\cdot)$ denotes the smooth-$L_1$ loss.

To encourage the ordering induced by the proxy targets, we use the pairwise ranking term
\begin{equation}
\begin{aligned}
\mathcal{L}_{\mathrm{rank}}
={}&\frac{1}{N}\sum_{i=1}^{N}
\frac{1}{\max\{1,|\mathcal{P}_i|\}}
\sum_{(m,n)\in\mathcal{P}_i}
\\[-2pt]
&\quad \log\!\left(
1+\exp\!\left(\rho_i^{(n)}-\rho_i^{(m)}\right)
\right),
\end{aligned}
\end{equation}
where $\mathcal{P}_i=\left\{(m,n)\mid \bar{\rho}_i^{(m)}>\bar{\rho}_i^{(n)}\right\}$; the inner sum is zero when $\mathcal{P}_i$ is empty. The overall objective of the Reliability-Aware Branch is
\begin{comment}
    \begin{equation}
    \mathcal{L}_{\mathrm{aware}}
=
\mathcal{L}_{\mathrm{rel}}
+\lambda_{r}\mathcal{L}_{\mathrm{rank}},
\end{equation}
\end{comment}
$\mathcal{L}_{\mathrm{aware}}
=
\mathcal{L}_{\mathrm{rel}}
+\lambda_{r}\mathcal{L}_{\mathrm{rank}}$
where $\lambda_r$ is a balancing coefficient.

\begin{table*}
\centering
\small
\setlength{\tabcolsep}{2.6pt} % 调整列间距，数字越小间距越窄
\renewcommand{\arraystretch}{1.0} % 调整行高
\begin{tabular}{lcccccc|cccccc}
\toprule
& \multicolumn{6}{c|}{\textbf{CMU-MOSI}} & \multicolumn{6}{c}{\textbf{CMU-MOSEI}} \\
\cmidrule(lr){2-7} \cmidrule(lr){8-13}
\textbf{Model} & Acc-2 & F1 & Acc-5 & Acc-7 & MAE ($\downarrow$) & Corr 
               & Acc-2 & F1 & Acc-5 & Acc-7 & MAE ($\downarrow$) & Corr \\
\midrule
MISA& 70.33/71.49 & 70.00/71.28 & 33.08 & 29.85 & 1.085 & 0.524
         & 75.82/71.27 & 68.73/63.85 & 39.39 & 40.84 & 0.780 & 0.503 \\
S-MM& 69.26/70.51 & 67.54/66.60 & 34.67 & 29.55 & 1.070 & 0.512
         & 77.42/73.89 & 72.31/68.92 & 45.38 & 44.70 & 0.695 & 0.498 \\
MMIM& 67.06/69.14 & 64.04/66.65 & 33.77 & 31.30 & 1.077 & 0.507
         & 75.89/73.32 & 70.32/68.72 & 41.74 & 40.75 & 0.739 & 0.489 \\
CENET & 67.73/71.46 & 64.85/68.41 & 33.62 & 30.38 & 1.080 & 0.504
         & 77.34/74.67 & 74.08/70.68 & 47.83 & 47.18 & 0.685 & 0.535 \\
TFR-Net& 66.35/68.15 & 60.06/61.73 & 34.67 & 29.54 & 1.200 & 0.459
         & 77.23/73.62 & 71.99/68.80 & 34.67 & 46.83 & 0.697 & 0.489 \\
ALMT& 68.39/70.40 & 71.80/72.57 & 33.42 & 30.30 & 1.083 & 0.498
         & 77.54/76.64 & 78.03/77.14 & 41.64 & 40.92 & 0.674 & 0.481 \\
LNLN& 70.94/72.55 & 71.25/72.73 & 38.27 & 34.26 & 1.046 & 0.527
         & 78.19/76.30 & 79.95/77.77 & 46.17 & 45.42 & 0.692 & 0.530 \\
P-RMF
         & 71.53/72.81 & 71.69/72.93 & 38.50 & 34.19 & 1.038 & 0.525
         & 78.83/78.14 & 80.39/79.33 & 45.87 & 44.63 & 0.658 & 0.589 \\
DAR
         & 71.60/73.18 & 71.51/73.15 & 38.65 & 34.47 & 1.069 & 0.520
         & 77.48/78.14 & 77.44/77.51 & 48.01 & 47.02 & 0.666 & 0.583 \\

MIG-HCL
         & 72.62 & 72.17 & - & 34.4 & 0.993 & 0.508 
         & 77.73 & 76.72 & - & 48.32 & 0.646 & 0.557 \\
\midrule
\textbf{MRCF} & \textbf{73.17/74.26} & \textbf{73.31/74.39} & \textbf{39.87} & \textbf{35.79} & \textbf{0.986} & \textbf{0.552}
         & \textbf{79.38/80.07} & \textbf{80.27/80.39} & \textbf{48.87} & \textbf{48.66} & \textbf{0.642} & \textbf{0.606} \\
\bottomrule
\end{tabular}
\caption{Results on CMU-MOSI and CMU-MOSEI under intra-modal missingness (IMM). Following the original evaluation protocol, results are averaged over the ten-rate grid $r\in\{0.0,0.1,\ldots,0.9\}$. For Acc-2 and F1, $a/b$ denotes the negative-versus-non-negative and negative-versus-positive label partitions, respectively.}
\label{table:CMU-IMM}
\end{table*}

\subsection{Reliability-Guided Interaction Branch}

Estimating reliability alone is not sufficient. The crucial question is how to use reliability once it has been estimated. This motivates our Reliability-Guided Interaction Branch.

The need for this design follows from the second phenomenon discussed in the introduction. When representations from low-reliability modalities enter cross-modal interaction without explicit reweighting, their distorted evidence is no longer confined to their original streams. Instead, its errors may propagate into other modalities and affect the joint representation. Therefore, reliability should not be used only to assign a higher or lower fusion weight at the end. It should also affect the earlier stage at which cross-modal messages are generated and routed.

For each ordered modality pair $(m,n)$ with $m\neq n$, we define a cross-modal message from source modality $n$ to target modality $m$. We first compute $Q_i^{(m)}=\mathbf{H}_i^{(m)}W_Q^{(m)}, K_i^{(n)}=\mathbf{H}_i^{(n)}W_K^{(n)}, V_i^{(n)}=\mathbf{H}_i^{(n)}W_V^{(n)}$, where $W_Q^{(m)},W_K^{(n)},W_V^{(n)}\in\mathbb{R}^{d\times d}$ are learnable projection matrices. The pairwise attention matrix is then
\begin{comment}
\begin{equation}
    A_i^{m\leftarrow n}
=
\mathrm{Softmax}\!\left(
\frac{Q_i^{(m)}(K_i^{(n)})^\top}{\sqrt{d}}
\right),
\end{equation}    
\end{comment}
$A_i^{m\leftarrow n}
=
\mathrm{Softmax}\!\left(
\frac{Q_i^{(m)}(K_i^{(n)})^\top}{\sqrt{d}}
\right)$
and the corresponding message is $\mathbf{U}_i^{m\leftarrow n}
=
A_i^{m\leftarrow n}V_i^{(n)}$, where $\mathrm{Softmax}(\cdot)$ is applied row-wise, so that $A_i^{m\leftarrow n}\in\mathbb{R}^{L_i^{(m)}\times L_i^{(n)}}$ and $\mathbf{U}_i^{m\leftarrow n}\in\mathbb{R}^{L_i^{(m)}\times d}$. 

A conventional interaction module would directly aggregate these pairwise messages. In contrast, MRCF explicitly regulates them according to modality reliability. We define the source-side routing weight as
\begin{comment}
    \begin{equation}
    \alpha_i^{m\leftarrow n}
=
\frac{
\exp\!\left(\rho_i^{(n)}\right)
}{
\sum\limits_{\substack{j\in\mathcal{M}\\ j\neq m}}
\exp\!\left(\rho_i^{(j)}\right)
}.
\end{equation}
\end{comment}
$  \alpha_i^{m\leftarrow n}
=
\frac{
\exp\!\left(\rho_i^{(n)}\right)
}{
\sum\limits_{\substack{j\in\mathcal{M}\\ j\neq m}}
\exp\!\left(\rho_i^{(j)}\right)
}.$
The updated representation of modality $m$ is then computed by
\begin{comment}
\begin{equation}
    \widetilde{\mathbf{H}}_i^{(m)}
=
\mathbf{H}_i^{(m)}
+
(1-\rho_i^{(m)})
\sum_{\substack{n\in\mathcal{M}\\ n\neq m}}
\alpha_i^{m\leftarrow n}\mathbf{U}_i^{m\leftarrow n}.
\end{equation}    
\end{comment}
$
    \widetilde{\mathbf{H}}_i^{(m)}
=
\mathbf{H}_i^{(m)}
+
(1-\rho_i^{(m)})
\sum_{\substack{n\in\mathcal{M}\\ n\neq m}}
\alpha_i^{m\leftarrow n}\mathbf{U}_i^{m\leftarrow n}$.

This design reflects the intended interaction logic. If target modality $m$ has a high reliability score, then $1-\rho_i^{(m)}$ is small and its update is reduced. If target modality $m$ has a lower score, the update is larger, with relatively more weight assigned to higher-scoring source modalities. Thus, reliability affects both the direction and magnitude of cross-modal information flow.

The proposed interaction mechanism is intended to preserve high-scoring modalities rather than overwrite them unnecessarily. It also turns cross-modal interaction into a reliability-aware process rather than uniform semantic mixing. To reduce the energy of outgoing messages from low-scoring source modalities, we use the following regularizer:
\begin{comment}
\begin{equation}
    \mathcal{L}_{\mathrm{prop}}
=
\frac{1}{N|\mathcal{M}|(|\mathcal{M}|-1)}
\sum_{i=1}^{N}
\sum_{\substack{m,n\in\mathcal{M}\\ m\neq n}}
\left(1-\bar{\rho}_i^{(n)}\right)
\left\|
\alpha_i^{m\leftarrow n}\mathbf{U}_i^{m\leftarrow n}
\right\|_F^2
\end{equation}    
\end{comment}

\begin{equation}
\begin{aligned}
\mathcal{L}_{\mathrm{prop}}
={}&
\frac{1}{N|\mathcal{M}|(|\mathcal{M}|-1)}
\sum_{i=1}^{N}
\sum_{\substack{m,n\in\mathcal{M}\\ m\neq n}}
\\[-2pt]
&\quad
\left(1-\bar{\rho}_i^{(n)}\right)
\left\|
\alpha_i^{m\leftarrow n}
\mathbf{U}_i^{m\leftarrow n}
\right\|_F^2 .
\end{aligned}
\end{equation}

This term penalizes the squared Frobenius norm of routed messages more strongly when the source has a low proxy-reliability score; it is therefore a reliability-weighted message-energy regularizer.

\subsection{Reliability-Calibrated Fusion Module}
\label{sec:rcfm}

After reliability-guided interaction, the model still needs to produce a final joint representation for sentiment prediction. A standard fusion strategy would pool the updated modality features and weight them through a learned attention mechanism. However, this strategy does not explicitly condition its modality weights on the estimated reliability scores.

Fusion under incomplete observations should satisfy two requirements simultaneously. It should preserve modality complementarity, because a modality with lower reliability may still retain partial but useful affective cues. It should also reduce the influence of low-reliability evidence on the final decision. A semantic-only attention mechanism does not explicitly address the second requirement. This motivates the Reliability-Calibrated Fusion Module.

We first pool each updated modality sequence $\widetilde{\mathbf{h}}_i^{(m)}
=
\mathrm{Pool}\!\left(\widetilde{\mathbf{H}}_i^{(m)}\right)$. The fusion weight of modality $m$ is then defined by jointly considering its semantic representation and estimated reliability:
\begin{equation}
    \beta_i^{(m)}
=
\frac{
\exp\!\left(
\mathbf{w}_f^\top[\widetilde{\mathbf{h}}_i^{(m)};\rho_i^{(m)}]
\right)
}{
\sum\limits_{n\in\mathcal{M}}
\exp\!\left(
\mathbf{w}_f^\top[\widetilde{\mathbf{h}}_i^{(n)};\rho_i^{(n)}]
\right)
},
\end{equation}
where $\mathbf{w}_f\in\mathbb{R}^{d+1}$ is a learnable parameter vector. Unlike a semantic-only fusion logit, the proposed logit also receives the estimated reliability score. The final contribution of a modality can therefore depend jointly on its semantic evidence and its estimated reliability under the current incomplete observation. In this way, reliability can influence the final representation rather than only an auxiliary loss.

The final fused representation is $\mathbf{z}_i
=
\sum_{m\in\mathcal{M}}
\beta_i^{(m)}\widetilde{\mathbf{h}}_i^{(m)}$, and the sentiment prediction is $\hat{y}_i=g_f(\mathbf{z}_i)$, where $g_f(\cdot)$ is the prediction head. To further stabilize fusion under incomplete observations, we introduce a complete-observation fusion anchor during training. Using the complete-observation features $\{\mathbf{h}_{i,c}^{(m)}\}_{m\in\mathcal{M}}$, we obtain $\mathbf{z}_i^{c}
=
\sum_{m\in\mathcal{M}}
\beta_{i,c}^{(m)}\mathbf{h}_{i,c}^{(m)}$, where $\beta_{i,c}^{(m)}$ is computed by the same fusion head. The corresponding complete-observation prediction is $\hat{y}_i^{c}=g_f(\mathbf{z}_i^{c})$. We then calibrate the incomplete-observation fused representation toward this complete-observation semantic anchor:
\begin{comment}
\begin{equation}
    \mathcal{L}_{\mathrm{cal}}
=
\frac{1}{N}
\sum_{i=1}^{N}
\left\|
\mathbf{z}_i-\operatorname{sg}\!\left(\mathbf{z}_i^{c}\right)
\right\|_2^2.
\end{equation}
\end{comment}
$    \mathcal{L}_{\mathrm{cal}}
=
\frac{1}{N}
\sum_{i=1}^{N}
\left\|
\mathbf{z}_i-\operatorname{sg}\!\left(\mathbf{z}_i^{c}\right)
\right\|_2^2$.

This calibration serves two purposes. It is intended to discourage the fusion module from over-pruning low-reliability modalities and discarding useful complementary evidence. It also discourages excessive deviation between paired incomplete- and complete-observation fused representations, thereby encouraging representation stability.

\subsection{Training Objective}

The main sentiment regression objective is
\begin{comment}
\begin{equation}
    \mathcal{L}_{\mathrm{task}}
=
\frac{1}{N}
\sum_{i=1}^{N}
\left|\hat{y}_i-y_i\right|.
\end{equation}    
\end{comment}
$    \mathcal{L}_{\mathrm{task}}
=
\frac{1}{N}
\sum_{i=1}^{N}
\left|\hat{y}_i-y_i\right|.$
The final objective of MRCF is
\begin{comment}
\begin{equation}
    \mathcal{L}
=
\mathcal{L}_{\mathrm{task}}
+\lambda_{1}\mathcal{L}_{\mathrm{aware}}
+\lambda_{2}\mathcal{L}_{\mathrm{prop}}
+\lambda_{3}\mathcal{L}_{\mathrm{cal}},
\end{equation}    
\end{comment}
$    \mathcal{L}
=
\mathcal{L}_{\mathrm{task}}
+\lambda_{1}\mathcal{L}_{\mathrm{aware}}
+\lambda_{2}\mathcal{L}_{\mathrm{prop}}
+\lambda_{3}\mathcal{L}_{\mathrm{cal}},$
where $\lambda_1,\lambda_2,\lambda_3$ are hyperparameters. During inference, MRCF operates on the available observations and does not require the complete-observation branch. The complete-observation branch and proxy reliability targets are used only during training.

\section{Experiments}

\subsection{Experimental Settings}

\textbf{Benchmarks.} We evaluate MRCF on widely standard MSA benchmarks, namely MOSI~\cite{Cmu-mosi}, MOSEI~\cite{Cmu-mosei}, and CH-SIMS~\cite{sims}. \textbf{Evaluation Metrics.} We evaluate MOSI and MOSEI using MAE, Corr, F1, Acc-2, Acc-5, and Acc-7, with Acc-2 and F1 reported under both negative–non-negative and negative–positive partitions, and CH-SIMS using MAE, Corr, F1, Acc-2, Acc-3, and Acc-5. \textbf{Implementation Details.} All experiments are conducted in PyTorch on an NVIDIA A100 GPU. We train for 100 epochs using Adam with a learning rate of $10^{-4}$ and a batch size of 32. BERT~\cite{BERT} is used as the language backbone, and the audio and visual feature extraction pipelines follow established baselines~\cite{ROSA,PRMF}. All reported results are averaged over three random seeds. Detailed provided in Appendix B. \textbf{Baseline references.} MISA~\cite{misa}, S-MM~\cite{self-mm}, MMIM~\cite{MMIM}, CENet~\cite{CENET}, TFR-Net~\cite{TFR-Net}, ALMT~\cite{ALMT}, LNLN~\cite{LNLN}, P-RMF~\cite{PRMF}, DAR~\cite{DRA}, MIG-HCL~\cite{MIG-HCL}, DCCAE(DE)~\cite{DCCAE}, DMD~\cite{dmd}, NIAT~\cite{NIAT}, GCNet~\cite{GCNet}, IMDer~\cite{IMDer}, EMT~\cite{EMT}, GMD~\cite{GGMD}, CoKD~\cite{CorrKD}, and ROSA~\cite{ROSA}.

% Detailed parameters are provided in Table~\ref{table:params}.

\subsection{Training Protocol}
We follow the two protocols used in prior incomplete-observation MSA~\cite{ROSA,PRMF}. Under intra-modal missingness (IMM), we erase a fraction $r\in\{0.0,0.1,\ldots,0.9\}$ of temporal positions. Missing-aware methods, including MRCF, P-RMF, LNLN, and ROSA, are trained and evaluated at the same target missing rate; methods requiring complete training inputs, such as MISA and S-MM, are trained on complete observations. All methods are evaluated with the same test corruption masks and operator at each $r$, and the main evaluation protocol reports mean performance across all ten missing rates. Under fixed-modality missingness (FMM), all methods are trained on complete observations and evaluated with one or two modalities unavailable. MRCF's paired complete-observation branch is used only during training and is absent at inference. Further corruption and input-handling details in Appendix B.

\subsection{Results under Intra-Modal Missingness}

We evaluate robustness to random temporal erasure on CMU-MOSI, CMU-MOSEI, and CH-SIMS. As shown in Tables~\ref{table:CMU-IMM} and~\ref{table:sims_IMM}, MRCF achieves the best or tied-best results on CMU-MOSI and CMU-MOSEI, while reaching 75.16\% Acc-2 and 79.71\% F1 on CH-SIMS, substantially outperforming P-RMF. It also obtains Corr scores of 0.552 and 0.606 on CMU-MOSI and CMU-MOSEI, respectively, demonstrating stronger consistency with continuous ground-truth sentiment under temporal corruption. Per-rate results are reported in Appendix B. Random sequence erasure disrupts local temporal continuity and induces reliability mismatch across modalities. Conventional fusion mechanisms may integrate corrupted streams without explicitly accounting for their unequal reliability, leading to reliability propagation bias. MRCF is designed to mitigate this issue. The Reliability-Aware Branch evaluates the internal quality of corrupted sequences and their cross-modal consistency. Guided by these estimates, the Reliability-Guided Interaction Branch reweights source messages and scales target-side updates according to estimated reliability. The resulting interaction is intended to reduce the influence of degraded streams at high missing rates.

\begin{table}[ht]
\centering
\setlength{\tabcolsep}{3pt}
\renewcommand{\arraystretch}{0.8}
\begin{tabular}{lcccccc}
\toprule
\textbf{Model} & \textbf{Acc-2} & \textbf{F1} & \textbf{Acc-3} & \textbf{Acc-5} & \textbf{MAE} ($\downarrow$) & \textbf{Corr} \\
\midrule
MISA& 72.71 & 66.30 & 56.87 & 31.53 & 0.539 & 0.348 \\
S-MM& 72.81 & 68.43 & 56.75 & 32.28 & 0.508 & 0.376 \\
MMIM& 69.86 & 66.21 & 52.76 & 31.81 & 0.544 & 0.339 \\
CENET& 68.13 & 57.90 & 53.17 & 22.29 & 0.589 & 0.107 \\
TFR-Net& 68.13 & 58.70 & 52.89 & 26.52 & 0.661 & 0.169 \\
% ALMT& 71.85 & 76.21 & 56.47 & 34.16 & 0.509 & 0.372 \\
P-RMF& 73.64 & 74.65 & 54.75 & 34.83 & 0.500 & 0.414 \\
LNLN& 72.73 & 79.43 & 57.14 & 34.64 & 0.514 & 0.397 \\
\midrule
\textbf{MRCF} & \textbf{75.16} & \textbf{79.71} & \textbf{57.97} & \textbf{35.44} & \textbf{0.481} & \textbf{0.435} \\
\bottomrule
\end{tabular}
\caption{Robustness comparison on SIMS under intra-modal missingness. Results are averaged over $r\in\{0.0,\ldots,0.9\}$.}
\label{table:sims_IMM}
\end{table}

\begin{table}[t] 
\centering
\small
\setlength{\tabcolsep}{2.0pt}
\renewcommand{\arraystretch}{0.8}
\begin{tabular}{c|cccccccc}
% \newcolumntype{M}{>{\centering\arraybackslash}m{1.5cm}}
\midrule
\multirow{2}{*}{\centering\textbf{Model}} & \multicolumn{8}{c}{\textbf{Available Modalities (F1, \%)}} \\
\cline{2-9}
& \rotatebox{90}{$\{l, v, a \}$}
& \rotatebox{90}{$\{l, v\}$}
& \rotatebox{90}{$\{l, a\}$}
& \rotatebox{90}{$\{v, a\}$}
& \rotatebox{90}{$\{l\}$}
& \rotatebox{90}{$\{v\}$}
& \rotatebox{90}{$\{a\}$}
& \rotatebox{90}{\textbf{Average}} \\
\midrule\midrule
\multicolumn{9}{c}{\textbf{CMU-MOSI}} \\
% \midrule
S-MM & 84.64 & 74.97 & 69.81 & 47.12 & 67.80 & 38.52 & 40.95 & 60.54 \\
DE& 66.76 & 65.43 & 62.25 & 42.39 & 61.78 & 41.45 & 41.30 & 54.48 \\
DMD & 84.50 & 68.45 & 70.51 & 50.47 & 68.97 & 42.26 & 43.33 & 61.21 \\
% MMIN~\cite{MMIN} & 85.20 & 84.76 & 83.50 & 45.51 & 82.39 & 43.86 & 44.25 & 67.07 \\
NIAT & 83.16 & 81.75 & 82.03 & 43.58 & 80.72 & 42.79 & 42.10 & 65.16 \\
GCNet & 83.20 & 83.58 & 84.73 & 70.02 & 81.12 & 59.67 & 68.07 & 75.77 \\
% CIF-MMIN & 84.27 & 82.48 & 81.62 & 54.70 & 81.16 & 51.09 & 52.02 & 69.62 \\
IMDer & 84.71 & 81.67 & 82.53 & 57.30 & 81.40 & 56.71 & 55.29 & 71.37 \\
EMT & 82.43 & 83.30 & 82.64 & 55.90 & 80.37 & 53.63 & 51.50 & 69.97 \\
GMD & 85.70 & 84.39 & 83.53 & 54.60 & 81.74 & 52.88 & 51.24 & 70.58 \\
CoKD & 83.94 & 82.41 & 82.36 & 73.74 & 81.20 & 60.72 & 66.52 & 75.84 \\
P-RMF & 84.37 & 81.94 & 82.10 & 73.11 & 81.36 & 70.32 & 71.44 & 77.81 \\
ROSA & 86.30 & 85.39 & 85.13 & 83.81 & 83.64 & 56.79 & 81.91 & 80.42 \\
\textbf{MRCF} & \textbf{86.79} & \textbf{86.08} & \textbf{85.85} & \textbf{84.47} & \textbf{84.34} & \textbf{61.67} & \textbf{82.35} & \textbf{81.65} \\
\midrule\midrule
\multicolumn{9}{c}{\textbf{CMU-MOSEI}} \\
% \midrule
S-MM& 83.69 & 74.62 & 75.91 & 49.52 & 71.53 & 37.61 & 43.57 & 62.35 \\
DE& 75.70 & 73.16 & 73.82 & 45.03 & 71.19 & 43.84 & 41.47 & 60.60 \\
DMD & 84.78 & 72.45 & 74.78 & 52.70 & 70.26 & 39.84 & 46.18 & 63.00 \\
% MMIN~\cite{MMIN} & 85.78 & 83.55 & 82.93 & 49.35 & 82.21 & 47.63 & 46.08 & 68.22 \\
NIAT & 86.35 & 85.31 & 85.10 & 48.57 & 83.29 & 45.76 & 44.43 & 68.40 \\
GCNet & 82.35 & 81.15 & 81.96 & 69.21 & 80.52 & 61.83 & 66.54 & 74.79 \\
% CIF-MMIN & 85.33 & 83.36 & 82.70 & 53.69 & 81.94 & 50.70 & 51.38 & 69.87 \\
IMDer & 84.71 & 82.39 & 83.41 & 53.30 & 82.77 & 51.83 & 50.47 & 69.84 \\
EMT & 85.17 & 84.45 & 83.81 & 52.39 & 83.04 & 50.79 & 51.28 & 70.13 \\
GMD & 87.07 & 85.52 & 83.10 & 52.67 & 81.45 & 49.96 & 50.39 & 70.02 \\
CoKD & 82.16 & 81.28 & 81.74 & 71.92 & 80.76 & 62.30 & 66.09 & 75.18 \\
P-RMF & 85.48 & 85.17 & 84.61 & 76.88 & 81.91 & 73.19 & 75.91 & 80.45 \\
ROSA & 89.56 & 87.32 & 86.64 & 84.19 & 84.42 & 54.38 & 79.04 & 80.79 \\
\textbf{MRCF} & \textbf{89.92} & \textbf{87.68} & \textbf{87.15} & \textbf{85.72} & \textbf{84.80} & \textbf{56.77} & \textbf{79.54} & \textbf{81.65} \\
\midrule
\end{tabular}
\caption{Results on CMU-MOSI and CMU-MOSEI under fixed-modality missingness (FMM). Here $l$, $v$, and $a$ denote the available language, visual, and acoustic modalities.} 
\label{table:FMM}
\end{table}

\subsection{Results under Fixed-Modality Missingness}

We further evaluate the framework under severe degradation, where one or two entire modalities are unavailable during inference. Table~\ref{table:FMM} presents performance under the fixed-modality-missingness conditions on CMU-MOSI and CMU-MOSEI. MRCF achieves the highest average F1 score of 81.65\% on both datasets. It also remains effective when the dominant language modality is removed. When only vision and audio are available, MRCF reaches an F1 score of 84.47\% on CMU-MOSI, exceeding the uncertainty-aware baseline ROSA by 0.66\%. Fixed-modality missingness removes entire semantic sources, making simple feature downweighting insufficient. Although existing methods may quantify uncertainty, constructing robust representations remains difficult when the primary semantic anchor is unavailable. MRCF addresses this representation gap through its Reliability-Calibrated Fusion Module. By anchoring incomplete-observation fusion to a complete-observation semantic target during training, the framework encourages the remaining modalities to retain generalizable affective cues. The Reliability-Guided Interaction Branch also enables reliability-dependent routing among the available modalities while reducing the influence of placeholder inputs. Together, these components contribute to robust sentiment prediction under severely incomplete inputs.

\begin{figure}[htbp]
    \centering
    \includegraphics[width=1\linewidth]{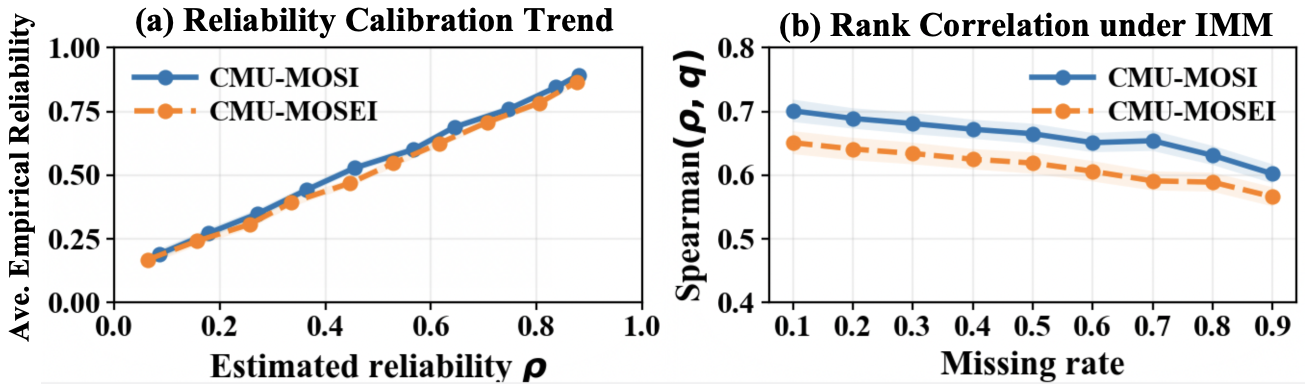}
  \caption{Alignment of modality reliability with a prediction-agreement diagnostic.
(a) Monotonic alignment on CMU-MOSI and CMU-MOSEI.(b) Spearman correlation across missing rates. Positive correlations indicate consistency with this model-derived diagnostic, not ground-truth reliability.}
\label{fig:1-REV}
\end{figure}

\subsection{Ablation Studies}
\label{sec:ablation}

Table~\ref{tab:ablation_mrcf_missingness} presents the ablation studies on CMU-MOSI, CMU-MOSEI, and CH-SIMS under intra-modal missingness. The reported F1 and MAE values are averaged over missing rates from 0.1 to 0.9. \textbf{Importance of Core Components.} Removing both the Reliability-Aware Branch and the Reliability-Guided Interaction Branch causes the largest component-level degradation (e.g., CMU-MOSI F1 decreases from 73.31\% to 69.90\%), suggesting that their contributions are complementary. Removing only the interaction branch causes a larger decrease than removing the Reliability-Aware Branch alone. This indicates that estimating reliability is most useful when the estimates also regulate cross-modal message passing. Omitting the Reliability-Calibrated Fusion Module also reduces performance, supporting the importance of reliability integration at the final decision stage. \textbf{Effectiveness of Reliability Estimation.} Within the estimation strategy, removing cross-modal consistency causes a larger decline than removing the intramodal quality cue. This result suggests that semantic compatibility across modalities is an important component of the estimated reliability under incomplete observations. Removing the ranking loss also reduces performance. \textbf{Interaction and Fusion Variants.} MRCF outperforms Uniform Interaction, indicating that reliability-aware message passing is useful relative to modality-agnostic interaction. Late-Reliability fusion also performs worse, supporting the use of reliability during interaction rather than only at final fusion. Addition and Concat+MLP yield lower results than the full model, further supporting the coupled interaction-fusion design. \textbf{Evaluation of Training Objectives.} Removing $\mathcal{L}_{\mathrm{rel}}$ causes the largest decrease among the loss-term ablations. Omitting $\mathcal{L}_{\mathrm{prop}}$ or $\mathcal{L}_{\mathrm{cal}}$ also consistently reduces performance, supporting their respective roles in message-energy regularization and complete-observation calibration.

\begin{table}[t]
\centering
\small
\setlength{\tabcolsep}{1.5pt}
\renewcommand{\arraystretch}{0.8}

\begin{tabular}{@{}lcccccc@{}}
\toprule
\multirow{2}{*}{\textbf{Variant}}
& \multicolumn{2}{c}{\textbf{CMU-MOSI}}
& \multicolumn{2}{c}{\textbf{CMU-MOSEI}}
& \multicolumn{2}{c}{\textbf{CH-SIMS}} \\
\cmidrule(lr){2-3}
\cmidrule(lr){4-5}
\cmidrule(l){6-7}
& \textbf{F1}$\uparrow$ & \textbf{MAE}$\downarrow$
& \textbf{F1}$\uparrow$ & \textbf{MAE}$\downarrow$
& \textbf{F1}$\uparrow$ & \textbf{MAE}$\downarrow$ \\
\midrule

\textbf{MRCF}
& \textbf{73.31} & \textbf{0.986}
& \textbf{80.27} & \textbf{0.642}
& \textbf{79.71} & \textbf{0.481} \\

\midrule
\multicolumn{7}{c}{\textbf{(1) Core Components}} \\
$w/o$ RAB  & 72.09 & 1.017 & 79.09 & 0.666 & 78.25 & 0.504 \\
$w/o$ RGIB & 71.86 & 1.024 & 78.95 & 0.672 & 77.96 & 0.510 \\
$w/o$ (R,G)& 69.90 & 1.063 & 77.42 & 0.694 & 76.35 & 0.535 \\
$w/o$ RCFM & 72.36 & 1.009 & 79.36 & 0.656 & 78.63 & 0.494 \\

\midrule
\multicolumn{7}{c}{\textbf{(2) Reliability Estimation}} \\
$w/o$ Intra-Q      & 72.46 & 1.001 & 79.56 & 0.654 & 78.86 & 0.492 \\
$w/o$ Cross-Cons.  & 71.98 & 1.015 & 79.02 & 0.668 & 78.15 & 0.506 \\
$w/o$ $\mathcal{L}_{\rm rank}$
                   & 72.76 & 0.997 & 79.74 & 0.650 & 79.02 & 0.489 \\

\midrule
\multicolumn{7}{c}{\textbf{(3) Interaction/Fusion Variants}} \\
Uniform Inter. & 71.22 & 1.033 & 78.53 & 0.676 & 77.54 & 0.514 \\
Late Rel.      & 71.59 & 1.026 & 78.74 & 0.670 & 77.81 & 0.509 \\
Addition       & 70.46 & 1.048 & 77.88 & 0.689 & 76.95 & 0.522 \\
Concat+MLP     & 70.85 & 1.041 & 78.10 & 0.684 & 77.13 & 0.517 \\

\midrule
\multicolumn{7}{c}{\textbf{(4) Training Objectives}} \\
$w/o$ $\mathcal{L}_{\rm rel}$
 & 71.95 & 1.015 & 78.96 & 0.668 & 78.01 & 0.507 \\
$w/o$ $\mathcal{L}_{\rm prop}$
 & 72.28 & 1.010 & 79.31 & 0.660 & 78.46 & 0.499 \\
$w/o$ $\mathcal{L}_{\rm cal}$
 & 72.62 & 0.999 & 79.58 & 0.651 & 78.87 & 0.492 \\
\bottomrule
\end{tabular}
\caption{Ablation studies of MRCF under intra-modal
missingness. $w/o$ (R,G) means $w/o$ (RAB+RGIB). F1 (\%).}
\label{tab:ablation_mrcf_missingness}
\end{table}

% \begin{table*}[ht]
% \centering
% \begin{tabular}{c c c c}
% \toprule
% \textbf{Hyper-parameter} & \textbf{CMU-MOSI~\cite{Cmu-mosi}} & \textbf{CMU-MOSEI~\cite{Cmu-mosei}} & \textbf{SIMS~\cite{sims}} \\
% \midrule
% $P, R$ & 256, 8 & 256, 8 & 256, 8  \\
% $\lambda_{\mathrm{CDA}}, \lambda_{\mathrm{KL}}$ & 0.1, 0.5 & 0.1, 0.5 & 0.1, 0.5  \\
% Batch size & 128 & 128 & 128  \\
% Epoch & 100 & 100 & 100  \\
% Optimizer & AdamW & AdamW & AdamW  \\
% Learning rate & $1\times 10^{-4}$ &$1\times 10^{-4}$& $1\times 10^{-4}$ \\
% Early Stop & $\checkmark$ & $\checkmark$ & $\checkmark$  \\
% Seed & 42,1111,1112,1113,1142& 42,1111,1112,1113,1142& 42,1111,1112,1113,1142\\
% \bottomrule
% \end{tabular}
% \caption{Hyper-Parameters Setting.}
% \label{table:params}
% \end{table*}

\subsection{Further Analysis}
\label{sec:further-analysis}

\textbf{Alignment with the Reliability Proxy.} We examine whether the estimated score is aligned with the intended prediction-agreement criterion. For sample $i$, modality $m$, and missing rate $r$, we define the diagnostic
\begin{comment}
    \begin{equation}
q_i^{(m)}(r)=\exp\!\left(-\left|\hat{y}_i^{(m,r)}-\hat{y}_i^{c}\right|\right),
\end{equation}
\end{comment}
$q_i^{(m)}(r)=\exp\!\left(-\left|\hat{y}_i^{(m,r)}-\hat{y}_i^{c}\right|\right),$
where $\hat{y}_i^{(m,r)}$ is the modality-$m$ prediction at missing rate $r$, and $\hat{y}_i^{c}$ is the complete-observation prediction defined in Sec.~\ref{sec:rcfm}. A larger $q_i^{(m)}(r)$ indicates closer prediction agreement. As shown in Fig.~\ref{fig:1-REV}, the estimated reliability $\rho_i^{(m)}$ is positively rank-correlated with $q_i^{(m)}(r)$ across CMU-MOSI and CMU-MOSEI. Furthermore, the same diagnostic shows that MRCF identifies the highest-scoring modality for most samples. Because $q_i^{(m)}(r)$ follows the same agreement principle as the training proxy, this analysis measures alignment with the intended criterion rather than independently validating actual modality trustworthiness. This result is also consistent with the ablation in Table~\ref{tab:ablation_mrcf_missingness}, where removing cross-modal consistency has a larger effect than removing the intramodal cue.

\begin{comment}
\begin{figure}[htbp]
    \centering
    \includegraphics[width=0.85\linewidth]{figures/2-RGIB.png}
  \caption{Reliability-guided cross-modal interaction.
(a) Interaction ratios across missing rates on CMU-MOSI, CMU-MOSEI, and CH-SIMS, relative to the baseline ($1.0$).
(b) CMU-MOSI routing at missing rates $0.2$ and $0.8$, showing stronger routing from higher- to lower-scoring modalities.}
\label{fig:2-RGIB}
\end{figure}    
\end{comment}
\begin{figure}[htbp]
    \centering
    \includegraphics[width=0.85\linewidth]{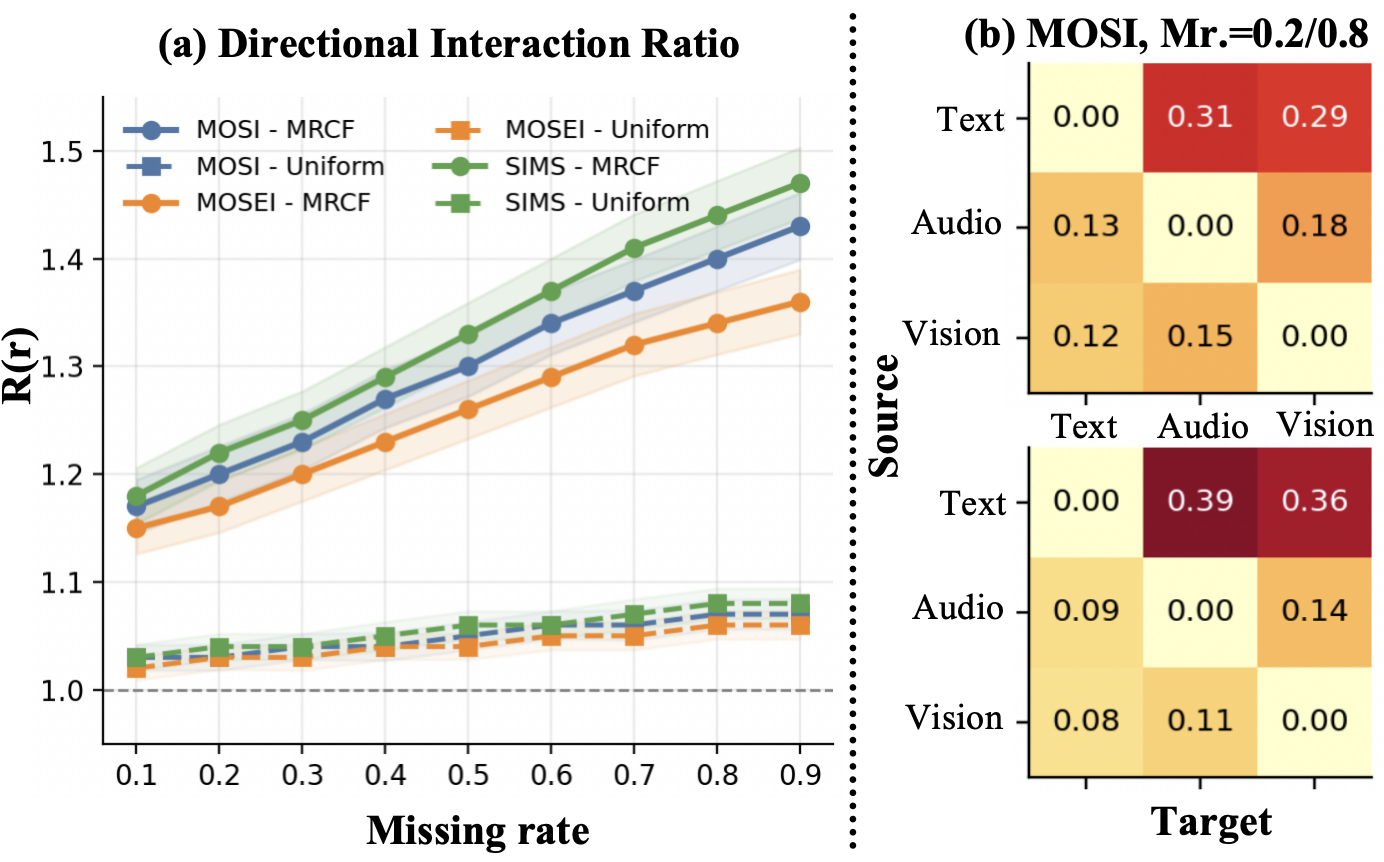}
 \caption{Reliability-guided cross-modal interaction.
(a) Interaction ratios across missing rates on three datasets.
(b) CMU-MOSI routing at $r=0.2$ and $0.8$.}
\label{fig:2-RGIB}
\end{figure}

\textbf{Behavior of Reliability-Guided Interaction.} To examine how estimated reliability affects cross-modal interaction, we define the effective message strength as $S_i^{m\leftarrow n}=(1-\rho_i^{(m)})\alpha_i^{m\leftarrow n}\|\mathbf{U}_i^{m\leftarrow n}\|_F$ and compute the ratio of interaction from higher- to lower-scoring modalities to interaction in the reverse direction. As shown in Fig.~4(a), MRCF consistently yields higher directional
interaction ratios than Uniform Interaction (near $1.0$), increasing
from $1.17$ to $1.43$ on MOSI as missingness grows. Figure~4(b)
further shows stronger routing from higher- to lower-scoring modalities
under severe missingness, supporting reliability-aware rather than
uniform semantic interaction.

\begin{figure}[ht]
    \centering
    \includegraphics[width=\linewidth]{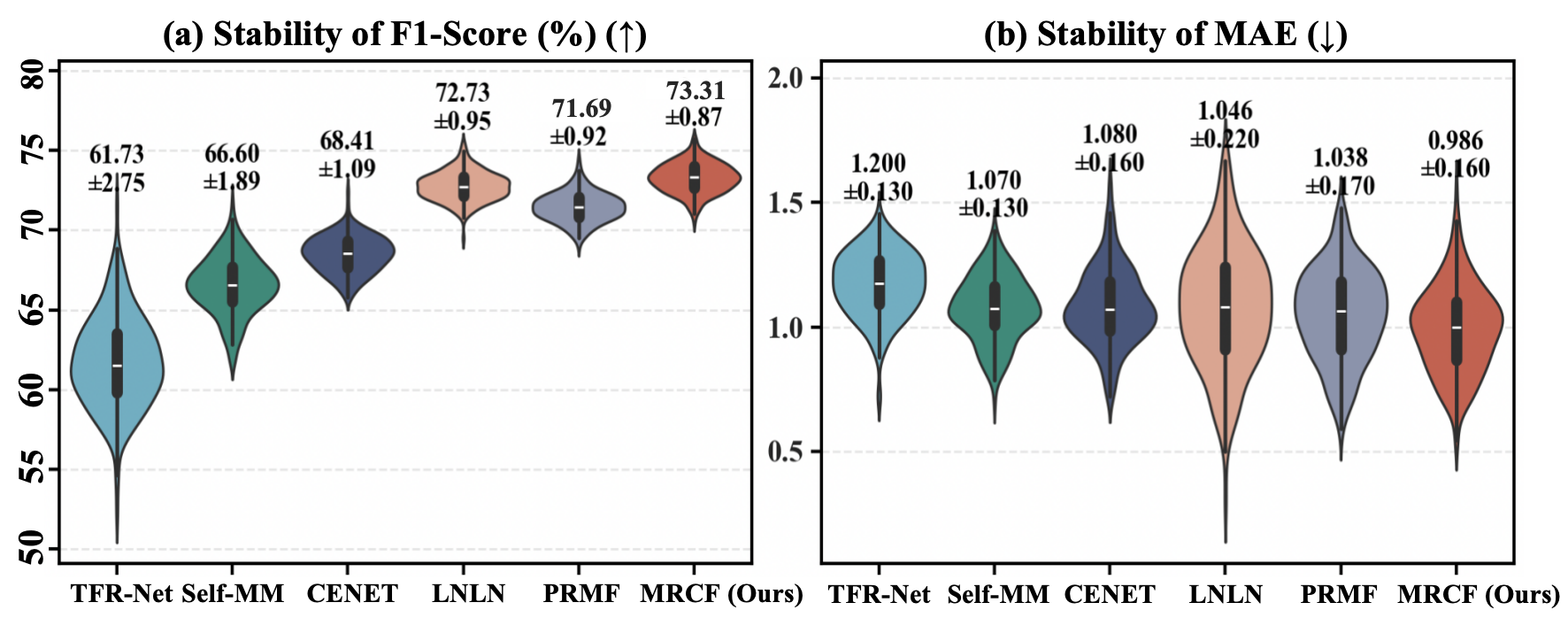}
    \caption{Statistical stability on MOSI. Plots show F1 and MAE distributions across missing rates and three seeds; box plots indicate the median and interquartile range (IQR).}
    \label{fig:Stability}
\end{figure}

\begin{comment}
\textbf{Statistical Stability Analysis.} We aggregate CMU-MOSI results across missing rates $r\in{0.0,0.1,\ldots,0.9}$ and three random seeds (Fig.~\ref{fig:Stability}). MRCF achieves a mean score of 73.31 and an MAE of 0.986, while exhibiting more concentrated distributions and narrower interquartile ranges than LNLN and P-RMF. These results demonstrate its stable performance under random modality missingness and support the effectiveness of reliability-aware interaction in suppressing degraded evidence.
\end{comment}
\textbf{Statistical Stability Analysis.}
Across missing rates and three seeds, MRCF achieves a mean F1 of
$73.31$ and an MAE of $0.986$, with tighter distributions than LNLN
and P-RMF (Fig.~5), indicating stable performance under random
modality missingness.

Branch-weight Visualizations, Hyperparameter Sensitivity and Efficiency Comparisons are in Appendix C, D, E.

\begin{comment}
\section{Conclusion}
    This paper investigates dynamically varying modality reliability and the resulting reliability mismatch and propagation bias in Multimodal Sentiment Analysis with incomplete observations. We propose MRCF, which explicitly models sample-specific modality reliability using intramodal quality and cross-modal consistency cues. The estimated reliability scores guide cross-modal message passing to suppress degraded evidence and are further incorporated into a reliability-calibrated fusion mechanism that aligns incomplete-observation representations with complete-observation anchors. Experiments on CMU-MOSI, CMU-MOSEI, and CH-SIMS demonstrate that MRCF achieves strong performance across diverse missingness settings, with additional stability and efficiency analyses supporting its robustness. These results highlight explicit reliability modeling as an effective principle for robust multimodal learning under incomplete observations.

\end{comment}
\section{Conclusion}
We study dynamic modality reliability in incomplete-observation MSA and
propose Modality Reliability-Calibrated Framework (MRCF) to address reliability mismatch and propagation bias.
MRCF estimates sample-specific reliability from intramodal quality and
cross-modal consistency, guides cross-modal interaction, and calibrates
fusion against complete-observation training anchors. Results on
CMU-MOSI, CMU-MOSEI, and CH-SIMS support explicit reliability modeling
across the evaluated missingness settings.

% \section{Conclusion}
% This paper studies dynamically varying modality reliability and the associated reliability mismatch and propagation bias in Multimodal Sentiment Analysis with incomplete observations. We propose MRCF, which models reliability as an explicit sample-specific variable rather than only as an implicit by-product of representation learning. MRCF estimates modality-reliability scores from intramodal quality and cross-modal consistency cues and is designed to address reliability mismatch. It then uses these scores to modulate cross-modal message passing and reduce the influence of degraded evidence, thereby mitigating reliability propagation bias. Coupled with a reliability-calibrated fusion mechanism that aligns incomplete-observation representations with complete-observation training anchors, the framework encourages more stable multimodal representations. Extensive evaluations on CMU-MOSI, CMU-MOSEI, and CH-SIMS show that MRCF achieves strong predictive performance across the evaluated missingness settings, while the stability and efficiency analyses provide complementary diagnostics. These findings suggest that explicit reliability modeling is a promising design principle for robust multimodal learning with incomplete observations.

\bibliography{aaai2027}


\end{document}